\documentclass[]{ceurart}
\pdfoutput=1

\usepackage{csquotes}

\usepackage{changes}
\definechangesauthor[name={Sirko}, color=orange]{s}
\definechangesauthor[name={Barbara}, color=blue]{b}

\usepackage[all]{nowidow}

\begin{document}

\copyrightyear{2021}
\copyrightclause{Copyright for this paper by its authors.
  Use permitted under Creative Commons License Attribution 4.0
  International (CC BY 4.0).}

\conference{S4BioDiv 2021: 3rd International Workshop on Semantics for Biodiversity, September 15, 2021, Bozen, Italy}

\title{The I-ADOPT Interoperability Framework for FAIRer data descriptions of biodiversity}

\author[1]{Barbara Magagna}[%
    orcid=0000-0003-2195-3997,
    email={barbara.magagna@umweltbundesamt.at},
]
\author[2]{Ilaria Rosati}[%
    orcid=0000-0003-3422-7230,
    email={ilaria.rosati@cnr.it},
]
\author[3]{Maria Stoica}[%
    orcid=0000-0002-6612-3439,
    email={maria.stoica@colorado.edu},
]
\author[4]{Sirko Schindler}[%
    orcid=0000-0002-0964-4457,
    email={sirko.schindler@dlr.de},
]
\author[5]{Gwenaelle Moncoiffe}[%
    orcid=0000-0001-6559-4178,
    email={gmon@bodc.ac.uk},
]
\author[6]{Anusuriya Devaraju}[%
    orcid=0000-0003-0870-3192,
    email={a.devaraju@uq.edu.au},
]
\author[1]{Johannes Peterseil}[%
    orcid=0000-0003-0631-8231,
    email={johannes.peterseil@umweltbundesamt.at},
]
\author[7]{Robert Huber}[%
    orcid=0000-0003-3000-0020,
    email={rhuber@uni-bremen.de},
]
\address[1]{Environment Agency Austria, Vienna, Austria}
\address[2]{Institute of Research on Terrestrial Ecosystems, National Research Council, Montelibretti, Rome, Italy}
\address[3]{University of Colorado, Boulder, Colorado}
\address[4]{Institute of Data Science, German Aerospace Center, Jena, Germany}
\address[5]{National Oceanography Centre, British Oceanographic Data Centre, Liverpool, United Kingdom}
\address[6]{Terrestrial Ecosystem Research Network (TERN), University of Queensland, Australia}
\address[7]{MARUM - Center for Marine Environmental Sciences, University of Bremen, Bremen, Germany}

\begin{abstract}
Biodiversity, the variation within and between species and ecosystems, is essential for human well-being and the equilibrium of the planet.
It is critical for the sustainable development of human society and is an important global challenge.
Biodiversity research has become increasingly data-intensive and it deals with heterogeneous and distributed data made available by global and regional initiatives, such as GBIF, ILTER, LifeWatch, BODC, PANGAEA, and TERN, that apply different data management practices. 
In particular, a variety of metadata and semantic resources have been produced by these initiatives to describe biodiversity observations, introducing interoperability issues across data management systems. 
To address these challenges, the InteroperAble Descriptions of Observable Property Terminology WG (I-ADOPT WG) was formed by a group of international terminology providers and data center managers in 2019 with the aim to build a common approach to describe what is observed, measured, calculated, or derived. 
Based on an extensive analysis of existing semantic representations of variables, the WG has recently published the I-ADOPT framework ontology to facilitate interoperability between existing semantic resources and support the provision of machine-readable variable descriptions whose components are mapped to FAIR vocabulary terms. 
The I-ADOPT framework ontology defines a set of high level semantic components that can be used to describe a variety of patterns commonly found in scientific observations. 
This contribution will focus on how the I-ADOPT framework can be applied to represent variables commonly used in the biodiversity domain. 
\end{abstract}


\maketitle

\section{Introduction}

The interactions and diversity of organisms within and across the Earth’s ecosystems play critical roles in the coevolution of the biosphere and the broader Earth system. 
Human societies have evolved with and are part of this complex system \cite{Folke2021}. 
Societal development and well-being depend on ecosystem services, or nature’s contributions to people \cite{Diaz2018}.
Global biodiversity, however, is declining faster than ever in human history \cite{Bongaarts2019}.
Biodiversity knowledge integration is urgently required to support responses towards its conservation and a sustainable future \cite{Heberling2021}.

Over the last decades, the need to share data and knowledge on biodiversity has led to numerous local, regional, and global initiatives resulting in an unprecedented biodiversity data mobilization \cite{Wueest2019}. 
Like many other domains, biodiversity research has been transformed by a big data revolution.
As the creation of large volumes of complex data becomes routine, new approaches using automated systems are required to find, access, combine, and interpret their meaning \cite{Wilkinson2016}.
The variety in data management practices and lack of common standards led to an increase of heterogeneity at multiple levels and severely impedes the integration of disparate data.
Semantic approaches promise to overcome this challenge by capturing rich representations of biodiversity data to facilitate maximum interoperability and provide detailed descriptions for re-use.
This has led to a large collection of independent terminology resources and tools across research infrastructures and communities as explored below.
Their complexity and diversity often overwhelm data managers and users, ironically maintaining barriers to interoperability.

LifeWatch Italy is one of the national branches of LifeWatch ERIC, the European e-science infrastructure for biodiversity and ecosystem research (LifeWatch ERIC\footnote{\url{https://www.lifewatch.eu/}}) \cite{Basset2012}. 
Coordinated by the National Research Council, LifeWatch Italy includes 35 partner institutions covering a wide range of scientific disciplines (terrestrial, marine and freshwater ecology, biology, zoology, botany, archaeobotany, geography, forestry, agriculture).
It seeks to reinforce integrated scientific research into biodiversity by the development of a National Hub as the main access point to data, apps, and eScience services for the management, aggregation, analysis, and re-use of biodiversity and ecosystem research data.
LifeWatch Italy has already developed a Data Management System which can receive biodiversity (e.g., checklists on geo-spatial distribution and abundance in Italy of vegetal and animal species) and ecosystem data (e.g., abiotic variables), including species traits data with grain size from individual to species (e.g., morphological traits of phytoplankton).
The system consists of two major components, the LifeWatch profile for the dataset description, based on the metadata standard EML 2.2.0\footnote{\url{https://eml.ecoinformatics.org/}} and a data schema based on the Darwin Core standard\footnote{\url{https://dwc.tdwg.org/}} and controlled vocabularies (e.g., LifeWatch Italy thesauri, EnvThes; in EcoPortal\footnote{\url{http://ecoportal.lifewatch.eu/}}), which provide a unified framework for the information management of LifeWatch Italy.
Thanks to a national infrastructure project, new developments and upgrades are already planned aiming specifically to improve the FAIRness of biodiversity and ecosystem data.

The emerging European Long-Term Ecosystem, critical zone and socio-ecological Research Infrastructure (eLTER RI\footnote{\url{https://www.lter-europe.net/elter-esfri}} \cite{Mirtl2018}) is currently in the preparatory phase building on LTER Europe, the regional network of ILTER. 
eLTER RI will adopt a fundamentally systemic approach to observe and analyze the environmental system, encompassing biological, geological, hydrological and socio-ecological perspectives. 
It will allow in-situ, co-located acquisition and gathering of Essential Variables ranging from bio-physicochemical to biodiversity and socio-ecological data. 
Ecosystem change caused by long-term pressures and short-term pulses will be investigated in a nested design across the scales covered by the eLTER RI site network. 
Biodiversity data for a number of species groups are collected by the contributing sites. \citeauthor{Pilotto2020} \cite{Pilotto2020} compiled 161 biodiversity time series mainly from European eLTER Sites focusing on freshwater ecosystems taking the effort to harmonize and standardize the data. 
A common infrastructure to harmonize species names still needs to be implemented. 
Currently, e.g. the R-package TaxonStand\footnote{\url{https://cran.r-project.org/web/packages/Taxonstand/}} is used to harmonize species data based on ThePlantList\footnote{\url{http://www.theplantlist.org/}}.
This will be further elaborated in the near future focusing on a harmonized reporting of biodiversity data. 
For reporting of species information a common data specification for eLTER \cite{Peterseil2020} has been developed following the recommendation of, e.g., the ICP Integrated Monitoring programme.\footnote{\url{https://unece.org/integrated-monitoring}}
For the future, biodiversity research and monitoring will be able to profit from a fully standardized and harmonized set of biodiversity standard observations. 
With this, many more sites could be used for various scientific analyses.
The same applies to a suitable set of environmental standard observations \cite{Zacharias2021} that help interpreting changes in biodiversity.

The British Oceanographic Data Centre (BODC) is managed and operated by the UK's National Oceanography Centre (NOC), an independent self-governing organization with a status of charitable company limited by guarantee.
BODC is the marine component of the Natural Environment Research Council's (NERC's) Environmental Data Centre network. 
Its mission is to develop, coordinate, and provide specialist data services for marine science communities; to enable innovative use and re-use of data; to ensure long-term curation of valuable and unique marine data resources; and to champion Open Data. 
BODC runs the NERC Vocabulary Server (NVS)\footnote{\url{ https://www.bodc.ac.uk/resources/products/web_services/vocab/}}, an internationally known and globally used infrastructure for the management of and access to controlled vocabularies related to the marine science and associated domains. 
Vocabularies served by the NVS underpin the SeaDataNet and EMODnet infrastructure. 
They have been adopted by the Ocean Biodiversity Information System (OBIS) to support data harmonization within and interoperability of the OBIS Darwin Core MeasurementOrFact Extension \cite{Pooter2017}.
These vocabularies, describing and identifying sensor and platform types and models, variables measured or derived, units, methods, and many other essential metadata elements,  are used by an increasing number of scientific marine data networks across the globe.

The information system PANGAEA\footnote{\url{https://www.pangaea.de/}} is jointly managed by the Alfred Wegener Institute Helmholtz Centre for Polar and Marine Research (AWI) and the Centre for Marine Environmental Sciences (MARUM) at the University of Bremen.
PANGAEA is a certified (CoreTrustSeal), trustworthy long-term operating repository providing continued access to more than 400,000 datasets from various sub-disciplines of Environmental Sciences. 
A large part of PANGAEA's dataset consists of biodiversity data, ranging from fossil records to modern faunal and floral observations collected through research infrastructures, research projects, and programs as well as individual researchers. 
The required observed variables are managed in a controlled vocabulary that is routinely undergoing an automated semantic annotation process \cite{Devaraju2021} to link, e.g., taxonomic vocabularies such as WORMS or ITIS to harmonize species-related data. 
Access to biodiversity datasets which are routinely tagged by a DOI is enabled through support for a number of community-specific and cross-domain standards. 

The Terrestrial Ecosystem Research Network (TERN)\footnote{\url{https://www.tern.org.au/}} collects and preserves biodiversity data together with other critical terrestrial ecosystem data from the continental scale to field sites across Australia to support researchers and policymakers' immediate and long-term needs, including state and regional governments. 
TERN has developed several valuable data services to support the discovery, analysis, and re-use of data. 
One of the core services is enabling data harmonization and integration through semantic data model\footnote{\url{https://ternaustralia.github.io/ontology_tern/}} (i.e., ontology) and machine-interpretable vocabularies. 
The TERN data model is developed based on the SOSA (Sensor, Observation, Sample, and Actuator)\footnote{\url{https://www.w3.org/TR/vocab-ssn/}} and comprises vocabularies for representing ecological survey data, including biodiversity, vegetation, and soil. 
The datasets are enriched with machine-interpretable vocabularies representing variables measured, features, units, methods, instruments, platforms, organization, and people. 
The vocabularies are available publicly through the TERN Linked Data Services\footnote{\url{https://linkeddata.tern.org.au/}}.
They are developed either in-house based on community-endorsed specifications (e.g., Australian Soil and Land Survey Field Handbook and AusPlots Rangelands Survey Protocols Manual) or imported from external semantic resources (e.g., QUDT\footnote{\url{https://www.qudt.org/}}, GCMD\footnote{\url{https://earthdata.nasa.gov/earth-observation-data/find-data/idn/gcmd-keywords}}, and NERC).
The extensible data model and rich collections of vocabularies are building blocks for improving TERN data curation and discovery.

These are just a few regional implementations demonstrating a highly diverse landscape of data management practices, even when dealing with the same type of resources. 
This makes it clear that there are still some difficult hurdles to overcome in order to achieve interoperability and convergence between infrastructures in the same domain, and even more challenging across domains. 
Initial progress has been made in providing machine-readable descriptions of sensors and their observation types through the OGC’s Sensor Web Enablement SensorML, Observations and Measurements (O\&M), the Sensor Observation Service (SOS) as well as the W3C’s/OGC’s Semantic Sensor Network (SSN) ontology \footnote{\url{https://www.ogc.org/docs/is}}.
This work has been further refined within the OGC SensorThings API specification, as well as the upcoming version 3 of O\&M.
However, \enquote{deep metadata} that further contextualizes observations (e.g., methodology or observable properties) is typically represented as coarsely qualified classes.
What exactly falls into these classes is currently unconstrained and could be anything ranging from unstandardized free-text to standardized descriptions accessible via fully resolvable URIs. 

To address the challenge to properly model and describe observable properties (also called variables), representative members of the above initiatives decided to collaborate under the umbrella of the Research Data Alliance. 
The Working Group InteroperAble Descriptions of Observable Property Terminology (RDA I-ADOPT WG)\footnote{\url{https://www.rd-alliance.org/groups/interoperable-descriptions-observable-property-terminology-wg-i-adopt}} was born in 2019 and is now in its final phase. 
This group of international terminology providers and data center managers set itself the objective to produce an Interoperability Framework paving the way for seamless terminology alignment for variable descriptions.
\section{Methodology}

This I-ADOPT WG has a strong focus on \textbf{variables observed in environmental research} because it leverages existing efforts to accurately encode what was measured, observed, derived, or computed in relation to Earth’s systems.
However, many of the principles it leans on are relevant to or connected with other domains. 
The construction of the framework has been informed by a review of current practices used in the community. 
The working group is also iteratively testing and refining the framework through a set of in-depth use cases.
Much like a generic blueprint, the refined conceptual framework will be a basis upon which terminology developers can formulate or refine their local design patterns to more easily align with others and to avoid insurmountable inconsistencies. 
By utilizing the recommendations of the I-ADOPT framework, terminology developers may expand the applicability and interconnectivity of their local resources in a collective attempt to uniformly represent complex properties observed across the environmental sciences, from marine, atmospheric, and terrestrial Earth sciences to biodiversity.

\begin{figure}[!htbp]
  \centering
  \includegraphics[width=\linewidth]{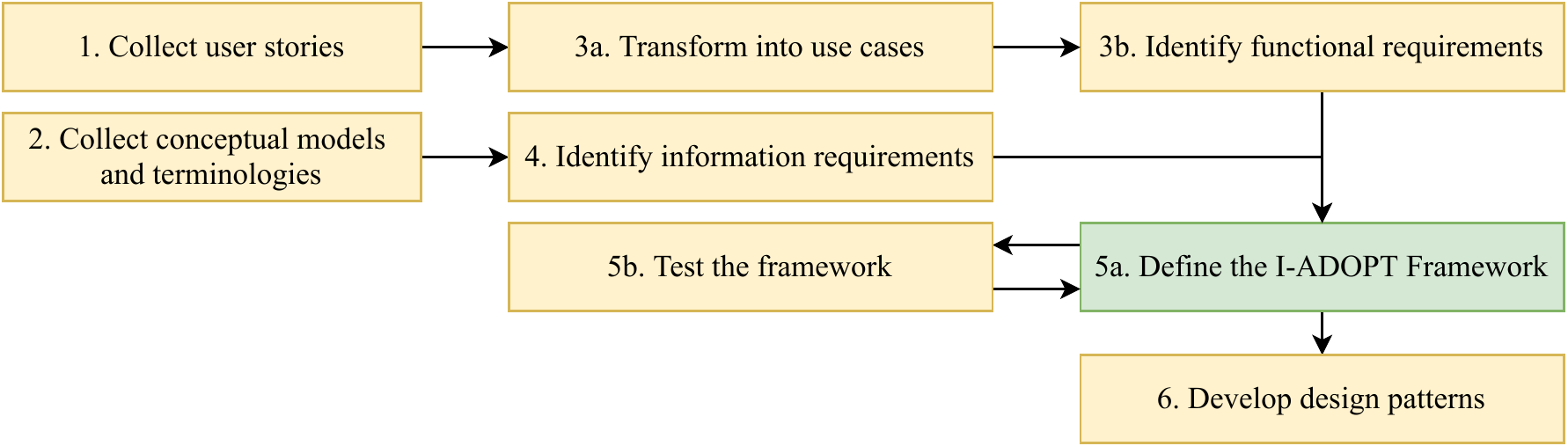}
  \caption{Tasks of the RDA WG I-ADOPT.}
  \label{fig:tasks}
\end{figure} 

The I-ADOPT WG has been working for about two years to develop the common framework for agreed-upon components to represent different aspects of observable properties. 
The timeline followed over the last few years involved completing six different tasks (see \autoref{fig:tasks}):
\begin{enumerate}
    \item Collect user stories from the community about needs related to semantic representation of observable properties involving nitrogen-related observable properties, following a submission template\footnote{\url{https://github.com/i-adopt/users_stories}}.
    \item Collect semantic representations from the community, including terminologies, standard vocabularies, ontologies, and conceptual models\footnote{\url{https://github.com/i-adopt/terminologies}}.
    \item Derive framework requirements by transforming user stories into concrete use cases, categorizing use cases, and defining functional needs within each category\footnote{\url{https://github.com/i-adopt/usecase_analysis}}.
    \item Analyze cataloged semantic artefacts to derive the framework's information requirements -- a set of conceptual classes and relationships common to the varying representation mechanisms, whether explicit or implicit.
    \item Develop the I-ADOPT Framework, including core variable components, their relationships and definitions, and iterate on feedback from the community.
    \item \textit{(ongoing)} Use the I-ADOPT Framework to develop aligned design patterns across several participating terminology resources for selected use cases.
\end{enumerate}

During the framework development, many potential framework components were considered, discussed, and iterated upon, but only a small subset of those are included in the current recommendation. 
Discussions of the different possible components that were in the end not considered essential to the framework, such as statistical measures, denominations of objects of interest, and property dimensions and units, are documented through the group's GitHub page and supplemental documents. These efforts resulted in an ontology for providing a machine-readable representation of the framework.

\section{Results}
\begin{figure}[!tbp]
  \centering
  \includegraphics[width=\linewidth]{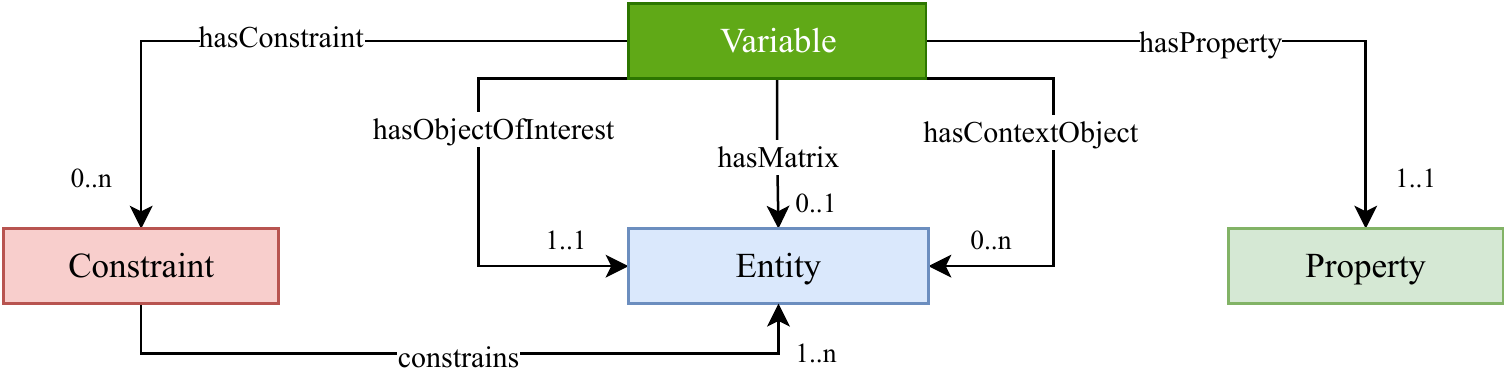}
  \caption{The I-ADOPT Framework.}
  \label{fig:framework}
\end{figure}

The I-ADOPT Framework ontology\footnote{\url{https://w3id.org/iadopt/ont/}} is a concise ontology (see \autoref{fig:framework}) made of four classes (Variable, Property, Entity, Constraint) and six object properties (hasProperty, hasObjectOfInterest, hasContextObject, hasMatrix, hasConstraint, constrains). It is designed to facilitate interoperability between existing variable description schemes (including domain-specific ontologies, semantic models, and structured controlled vocabularies) and to support further development of machine-readable variable descriptions that re-use components mapped to FAIR vocabulary terms. 
This first official version of the ontology has been developed by a core group of terminology experts and users from the I-ADOPT WG. 

The framework is variable-centric and can be used to describe result values of any type of data acquisition events, be it a human-based observation, a sensor-based measurement, a calculation, or a simulation. 
The Variable describes WHAT has been observed, measured, simulated, or calculated independently of WHERE (site description, geographical coordinates), HOW (procedure, protocol) and WHEN (measurement time, time resolution) the data acquisition has taken place, although in some cases the HOW might be important to include in the variable description, as it might pose certain constraints on it.
This makes the application of this concept reusable in different settings giving meaning to the determined value.
Variables exist as concepts in many vocabularies (see CF Standard Names\footnote{\url{https://cfconventions.org/Data/cf-standard-names/47/build/cf-standard-name-table.html}} or CSDMS Standard Names\footnote{\url{https://csdms.colorado.edu/wiki/CSDMS_Standard_Names}}) to provide metadata for the values provided in databases. 
They are also known as observable properties (see Observable Property Vocabulary\footnote{\url{http://registry.it.csiro.au/def/environment/_property}}) or parameters (see BODC Parameter Usage Vocabulary (P01)\footnote{\url{https://www.bodc.ac.uk/resources/vocabularies/parameter_codes/}} and EnvThes\footnote{\url{http://vocabs.lter-europe.net/EnvThes/}}). 
This variety of approaches and the lack of a common representation strategy lead to a situation where variable concepts from different terminologies are still not directly comparable. Consequently, the data that is annotated with these concepts is not interoperable and requires substantial manual efforts to integrate.

\begin{figure}[!tbp]
  \centering
  \includegraphics[width=\linewidth]{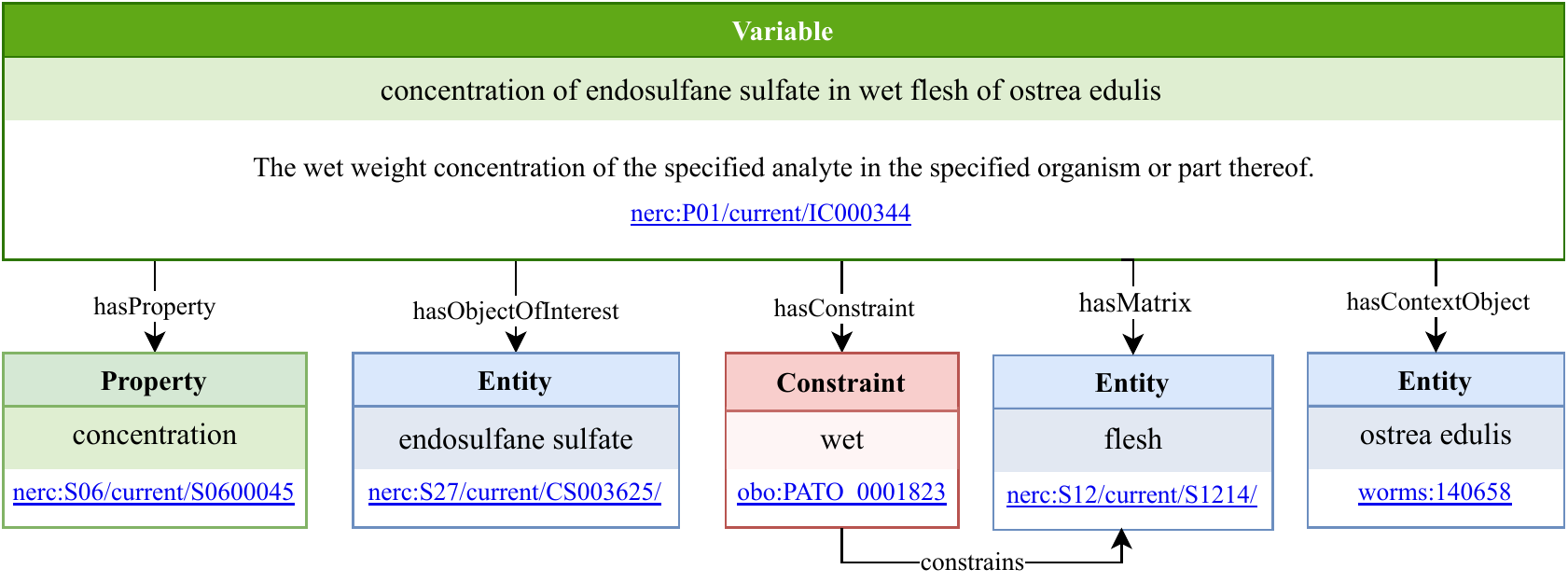}
  \caption{Quantitative example: Concentration of endosulfan sufalte in Ostrea edulis.}
  \label{fig:example_quan}
\end{figure}

In the following paragraphs, we explain the implementation of the framework by use of an example\footnote{
    The following prefixes will be used throughout the examples:
    \texttt{
        obo:<\url{http://purl.obolibrary.org/obo/}> ; \newline
        lifewatch:<\url{http://thesauri.lifewatchitaly.eu/PhytoTraits/index.php?}> ;\newline 
        nerc:<\url{http://vocab.nerc.ac.uk/collection/}> ;
        worms:<\url{http://marinespecies.org/aphia.php?p=taxdetails&id=}>
    }
}, describing a complex biodiversity variable that requires all of the components of the framework. 
The variable description is \enquote{concentration of endosulfan sulfate in wet flesh of ostrea edulis} (see \autoref{fig:example_quan}) and refers to the quantitative result (i.e., requiring a magnitude and unit) of a measurement. 

The I-ADOPT ontology, inspired by the atomisation approach of the Complex Property Model \cite{Leadbetter2015} and The Scientific Variables Ontology\footnote{\url{http://www.scientificvariablesontology.org/}} \cite{Stoica2018,Stoica2019}, conceives the \textbf{Variable} as a compound concept consisting of at least one entity (the ObjectOfInterest) and its Property, but very often includes further entities contextualizing the target object of observation.
Splitting the Variable into constituent concepts enables the reuse and the mapping of these components in the context of other variables. 
The ObjectOfInterest (\textit{endosulfan sulfate}) can be identified independently of an observable characteristic (\textit{concentration}).
By concatenating the constituent elements with prepositions the preferred label of a variable can be automatically created (compare with the right side of the table in \autoref{fig:example_quan}). 
However, the I-ADOPT Framework does not provide any recommendation how the label of the variable should be composed or which order to follow as its main contribution targets machine-readability of the concept. 
Not included in the variable description are units, measurement methods, and time-related or geographical location information.
Units are essential information for describing measures, but a quantitative variable might be expressed in different units which requires units be modeled independently of variables. 
By isolating the Variable concept from these supplementary information elements, it is possible to use the I-ADOPT framework in a broader context interoperating with other representations of observations.
In this way the Variable with its explicit constituents can be conceived as an extended and machine-readable conceptualization of the respective WHAT concept of these models (e.g. ObservableProperty in SOSA\footnote{\url{https://www.w3.org/TR/vocab-ssn/}}, Characteristic in OBOE\footnote{\url{https://github.com/NCEAS/oboe/}}).

The \textbf{Property} (\textit{concentration}) is a type of characteristic of the ObjectOfInterest. 
This characteristic inheres in the object and exists as long as the object exists.
The cardinality of the object property hasProperty is $1..1$, meaning that a Variable has to have exactly one Property. 
The Property is in most cases a more generalized concept than the variable itself, as the variable adds more detail to the property.
This has the advantage that the property term can be reused in other variable descriptions and further on allows to identify similar measurements across different variables (e.g., \textit{concentration} regardless of the substance observed).

The \textbf{Entity} is any object (\textit{endosulfan sulfate}) or process (e.g., erosion) that has a role in an observation.
An Entity may play one of the following roles: ObjectOfInterest, ContextObject, or Matrix. 
The roles of the entities in the variable description are implemented as object properties in the I-ADOPT ontology.
Whether the involvement of a particular entity is meaningful enough to include in the variable description, depends on the specific context.

\begin{itemize}
    \item \textbf{hasObjectOfInterest}: The ObjectOfInterest (\textit{endosulfan sulfate}) is the Entity whose property is observed. A Variable requires exactly one ObjectOfInterest (cardinality: $1..1$).
    \item \textbf{hasContextObject}: The ContextObject (\textit{Ostrea edulis}) is the Entity that provides additional background information regarding the ObjectOfInterest. A variable can have more than one ContextObject (cardinality: $0..n$).
    \item \textbf{hasMatrix}: The Matrix (\textit{flesh}) is the Entity in which the ObjectOfInterest is contained. A Variable might only have one Matrix (cardinality: $0..1$).
\end{itemize}

The \textbf{Constraint} (\textit{wet}) limits the scope of the observation and confines the context to a particular state.
It describes properties of the involved entities (\textit{flesh}) that are relevant to the particular observation. 
Oftentimes Constraints take the form of observations themselves, fixing other properties that influence the main observation considered. 
It may constrain (\textbf{constrains}) multiple entities playing the role of ObjectOfInterest, Matrix, or ContextObject.
Constraints are optional, so their the cardinality of the object property \textbf{hasConstraint} is $0..n$.

The I-ADOPT ontology provides a framework to adequately address the data interoperability requirements emphasized by the FAIR Guiding Principles \cite{Wilkinson2016}.
Appropriate terms from FAIR semantic artefacts (like SKOS vocabularies or OWL ontologies) should be used not only for the variable itself, but also for each of the components of the variable description.
This provides qualified references to other metadata (FAIR Principle I3) where these exist or can be easily created. 
In order to model a variable accurately, sound domain-specific knowledge is required.
This knowledge should be sufficiently reflected in the form of a human-readable definition of the variable concept. 
Applying the I-ADOPT framework at the metadata level means using the persistent and resolvable identifier of the variable concept (e.g., Internationalized Resource Identifier - IRI) instead of plain text information in the metadata field addressing the observed property (FAIR Principle I2).
This results in semantically rich, machine-readable descriptions (metadata) of the dataset. 
At the dataset level, the provision of FAIR data instead depends very much on the distribution format of the datasets. 
In the case of spreadsheets, where the quantitative values are described by the column headings, variable concept IRIs should be used instead of labels (see \autoref{fig:example_quan}).
In an ideal FAIR world, data providers would distribute their datasets in a formal, accessible, shared, and broadly applicable language for knowledge representation like RDF (Resource Description Language) (FAIR Principle I1) as a way describing resources on the web.
Using RDF allows to use IRIs of the variable concept directly in the triple statement describing the measurement result. 
\citeauthor{Leadbetter2015} \cite{Leadbetter2015} demonstrated that by FAIRifying metadata it is possible to produce RDF files of the semantically enriched datasets. 
Because the variable itself links to further descriptive elements, making implicit knowledge about the measurement explicit, it is possible to apply faceted semantic data search via SPARQL queries on data platforms, allowing the selection of datasets involving a substance (like \textit{endosulfan sulfate}) or a specific ObjectofInterest (like \textit{Ostrea edulis}).

\begin{figure}[!tbp]
  \centering
  \includegraphics[width=\linewidth]{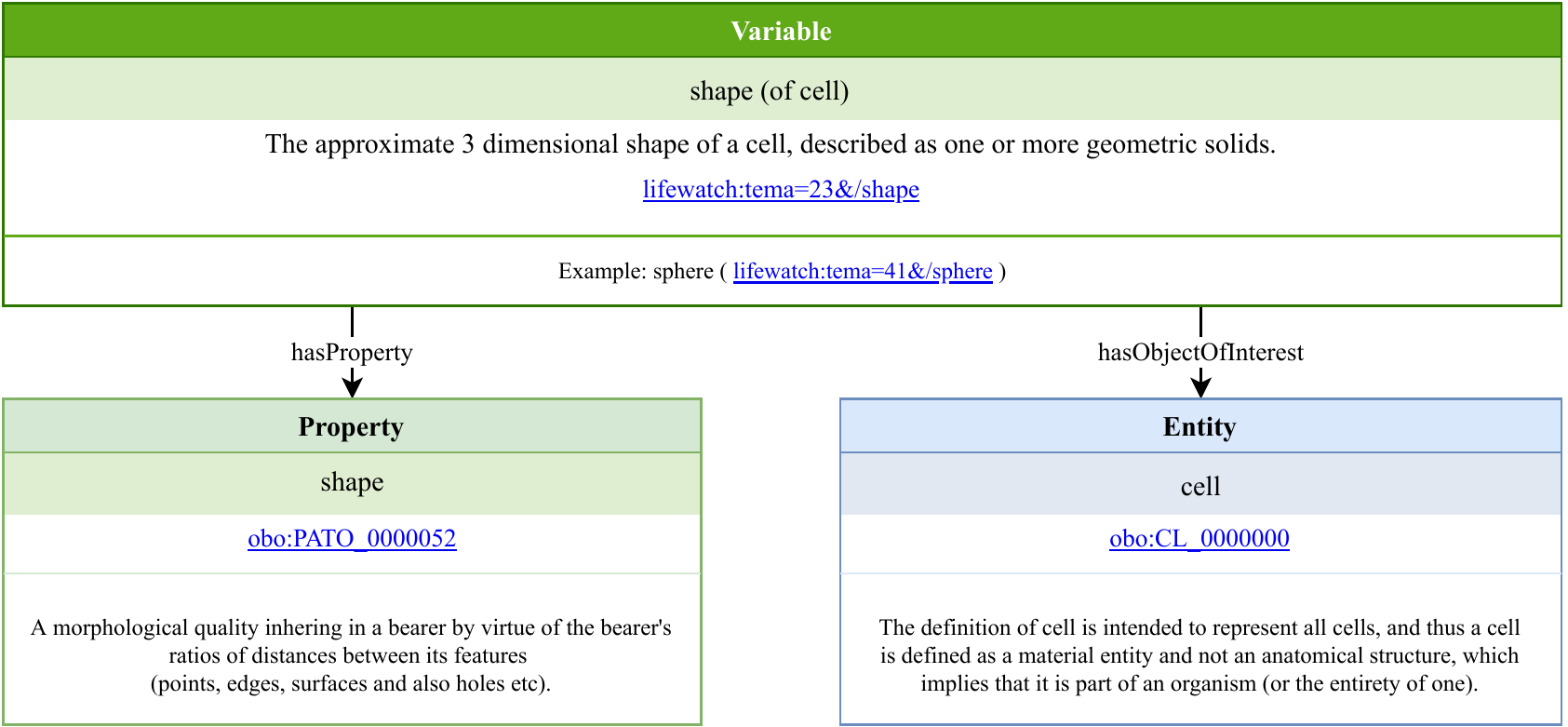}
  \caption{Qualitative example: Shape of cell.}
  \label{fig:example_qual}
\end{figure}

Many variables in the biodiversity domain are human-based observations with a qualitative result value selected from a controlled list. 
Some elements for the Occurrence, Organism, and Taxon metadata components of the Darwin Core standard require qualitative values for which a few recommended SKOS vocabularies were created by the TDWG Darwin Core Maintenance Group\footnote{\url{https://dwc.tdwg.org/list/}}.
LifeWatch Italy vocabularies provide specific terminology for functional traits of several groups of aquatic organism and alien species\cite{Rosati2017}.
An example of how the I-ADOPT framework can be applied to qualitative results of human-based observations is provided in \autoref{fig:example_qual} using a variable from the LifeWatch Phytotraits thesaurus. 
The variable is called \textit{shape} implicitly targeting cells, which is explained in the definition of this specific variable.
It is implemented as a SKOS concept and has as more specific concepts qualitative values such as \textit{sphere}, \textit{cube}, or \textit{ellipsoid}. 
In dataset spreadsheets these values would normally populate the cells of the variable column, but we recommend to use instead the IRIs of these narrower concepts as manifestations of the observed value.
However, a more convenient way to store this type of information would be to use RDF for the entire dataset. 
The constituent components of the variable \textit{shape (of cell)} are the more general concept of \textit{shape} (the \textbf{Property}) and the \textbf{ObjectOfInterest} the concept \textit{cell}.
It is not always possible to find more general properties which could be reused in the context of other variables. 
The Darwin Core metadata element \textit{degreeOfEstablishment} would be referred to as a variable with the IRI \footnote{\url{https://dwc.tdwg.org/doe/}} and decomposed into the Property \textit{degreeOfEstablishment} with the same IRI and the ObjectOfInterest \textit{organism}\footnote{\url{http://purl.obolibrary.org/obo/OBI_0100026}}.

\section{Outlook}

The RDA WG I-ADOPT will in its final phase produce guidelines for the community on how to apply the framework. 
These will include design patterns to generalize variable descriptions and their modelling for their reuse in different settings (e.g., \textit{concentration of substance in biota}). 
The authors of this paper will evaluate the adoption of the I-ADOPT approach in their own infrastructures. 
One  first proof-of-concept will be the SKOS representation of the eLTER Standard Observation variables based on the I-ADOPT Framework in EnvThes. 
An important task will also be to work on mappings to existing observation models (e.g., OBOE) and to foster the ongoing collaboration with initiatives like OGC Observations and measurements Standard Working Group\footnote{\url{https://github.com/opengeospatial/om-swg}} and DDI-CDI\footnote{\url{https://ddi-alliance.atlassian.net/wiki/spaces/DDI4/pages/860815393/DDI+Cross+Domain+Integration+DDI-CDI+Review}} to evaluate how I-ADOPT might increase semantic interoperability of the variable concept within those approaches. 

\begin{acknowledgments}
  The authors like to thank the Research Data Alliance, eLTERplus, ENVRI-FAIR (grant agreements No 871128, No 824068) and Alison Pamment (NCAS, UK) for their support.
\end{acknowledgments}
\vspace{-0.3em}

\bibliography{tex/references}

\end{document}